\documentclass[10pt,twocolumn,letterpaper]{article}

\usepackage{cvpr}
\usepackage{times}
\usepackage{epsfig}
\usepackage{graphicx}
\usepackage{amsmath}
\usepackage{amssymb}
\usepackage{multirow}
\usepackage{caption}
\usepackage{subfigure}
\usepackage{color}
\usepackage[table,xcdraw]{xcolor}
\usepackage{CJKutf8}

\cvprfinalcopy %

\begin{document}

\title{On the General Value of Evidence, and \\ Bilingual Scene-Text Visual Question Answering}

\author{
Xinyu Wang$ ^{1*} $, ~ ~ ~ Yuliang Liu$ ^{1,2} $\thanks{XW and YL contributed equally.
YL's contribution was made when visiting The University of Adelaide.},
~ ~ ~ Chunhua Shen$ ^1 $\thanks{Corresponding author, e-mail:
$\sf chunhua.shen@adelaide.edu.au$
}, ~ ~ ~ Chun Chet Ng$ ^3 $,
~ ~ ~ Canjie Luo$ ^2 $
\\
Lianwen Jin$ ^2 $, ~ ~~  Chee Seng Chan$ ^3 $,
~ ~~ Anton van den Hengel$ ^1 $, ~ ~ ~ Liangwei Wang$ ^4 $
\\
[0.2cm]
$ ^1 $ University of Adelaide
~
$ ^2 $ South China University of Technology
~
$ ^3 $ University of Malaya
~
$ ^4 $ Huawei
}

\maketitle
\begin{abstract}
Visual Question Answering (VQA) methods have made incredible progress, but suffer from a failure to generalize.  This is visible in the fact that they are vulnerable to learning coincidental correlations in the data rather than deeper relations between image content and ideas expressed in language.  We present a dataset that takes a step towards addressing this problem in that it contains questions expressed in two languages, and an evaluation process that co-opts a well understood image-based metric to reflect the method's ability to reason.  Measuring reasoning directly encourages generalization by penalizing answers that are coincidentally correct.  The dataset reflects the scene-text version of the VQA problem, and the reasoning evaluation can be seen as a text-based version of a referring expression challenge.  Experiments and analysis are provided that show the value of the dataset.

\end{abstract}

\section{Introduction}
The fact that Visual Questions Answering \cite{antol2015vqa} methods are able to answer natural language questions that relate to a wide variety of image content has been an incredible development. The limitations of existing methods, and particularly their tendency to focus on spurious correlations in the data, have been repeatedly identified (see~\cite{agrawal2018don,goyal2017making,clevr2017}, for example). This is visible in the tendency of methods to answer questions on the basis of the text alone. The answer to `How many' questions, for instance, is predominantly `Two'.

Focusing on coincidental correlations in the data represents a failure to generalize. These correlations are not stable across datasets, meaning that once the test data moves beyond the distribution of the training set, the correlations fail to hold, and methods that exploit them fail to work. The underlying reasoning, in contrast, is stable across datasets. Encouraging VQA methods to reason about the image content is thus critical to achieving methods that generalize.

One of the underlying problems with encouraging VQA methods to generalize has been that it is impossible to tell whether a method arrived at the right answer through for the right reasons. An answer is equally correct whether it results from analysis of the underlying reasoning or through exploiting a coincidental correlation in the data. A series of works have developed more sophisticated measures of performance for vision and language problems \cite{anderson2018vision,goyal2017making,yi2018neural}, and this work falls in this category. What distinguishes this approach is that it uses image-based grounding to encourage generalization, despite the fact that it is not actually required to achieve the desired task.

\begin{figure}[tb!]
    \centering
    \includegraphics[width=.35\textwidth]{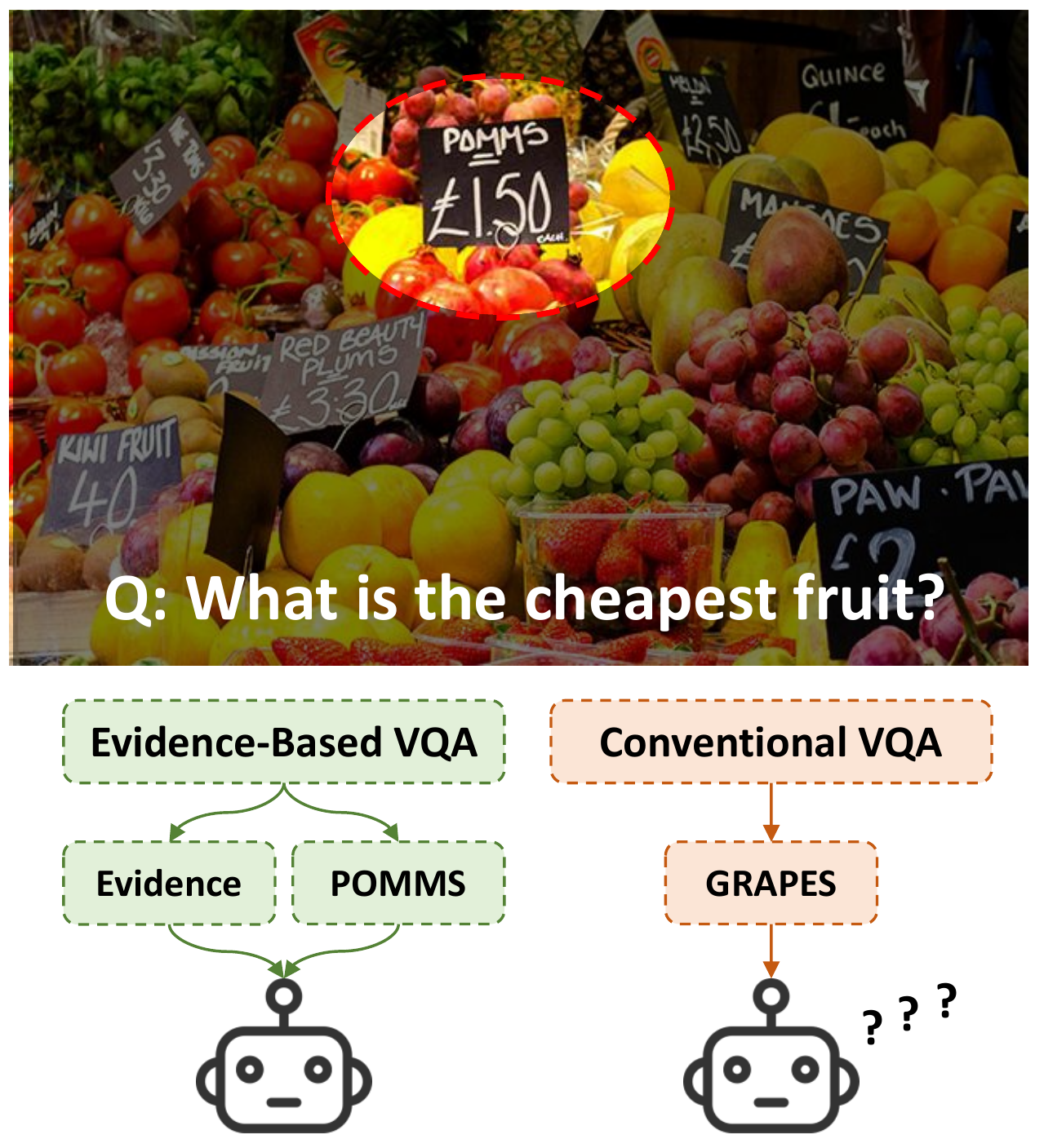}
    \caption{%
    Requiring that vision-and-language methods provide evidence for their decisions encourages the development of approaches that depend on reasoning, and thus that are better able to generalize to new situations.
    It also helps generate confidence in the provided answer.}
    \label{fig:background}
    \vspace{-0.5cm}
\end{figure}

We propose here an approach to measuring VQA performance that encourages generalization by demanding that the algorithm justify its reasoning (see Fig.~\ref{fig:background}). Previous methods have applied the same rational, but suffered because the form in which the reason must be provided is constrictive~\cite{Wang:2018, wu2016ask}. We show here that it is possible instead to evaluate reasoning by requiring only that a method provide a relatively simple indication of which area of the image it has based its answer on. If the method has provided the correct answer and the correct image region then it is likely that it has employed the right reasoning. Using image regions, or more accurately bounding boxes, as an evaluation metric also has the advantage that Intersection-over-Union (IoU) measures are well understood in the field.

\begin{figure*}[t!]
    \centering
    \subfigure[]{\includegraphics[width=3.5cm,height=3.2cm]{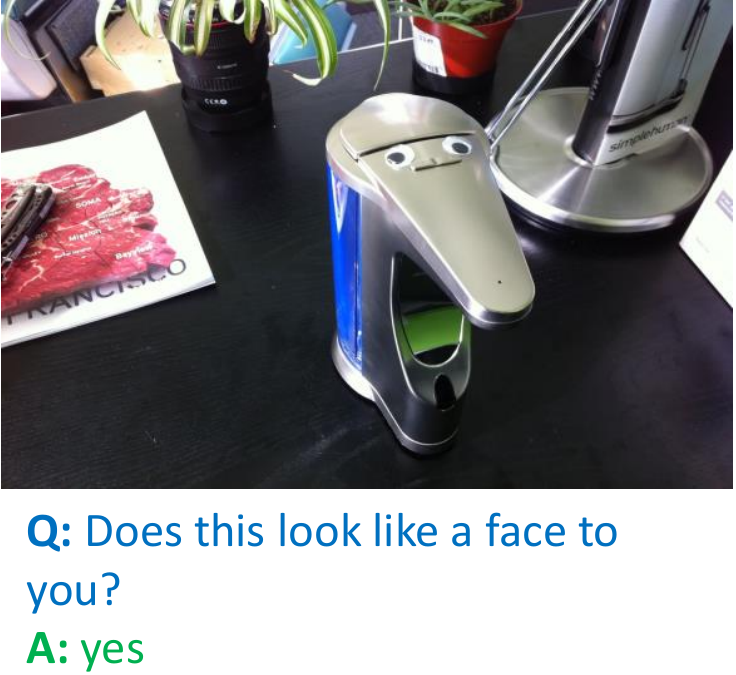}
    \label{fig:exist_a}}\hspace{0.05cm}
    \subfigure[]{\includegraphics[width=3.5cm,height=3.2cm]{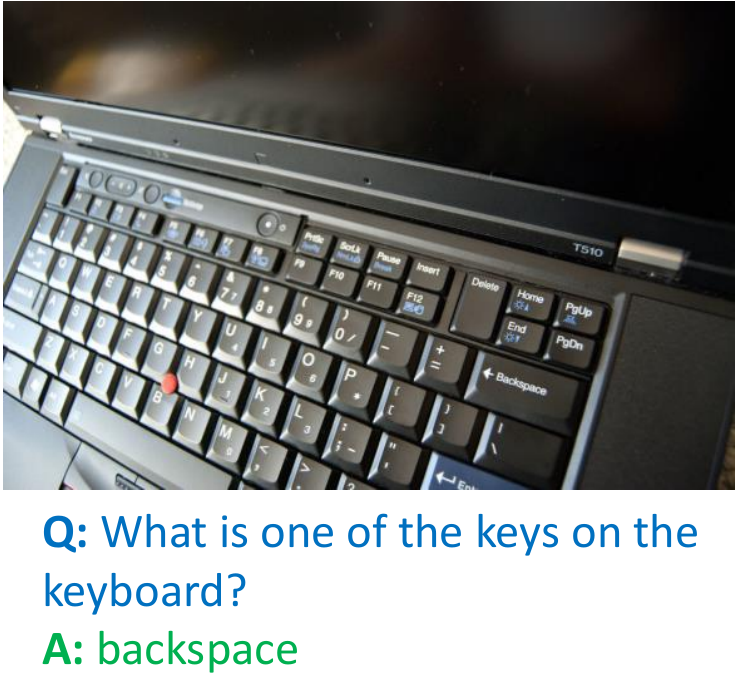}
    \label{fig:exist_b}}\hspace{0.05cm}
    \subfigure[]{\includegraphics[width=3.5cm,height=3.2cm]{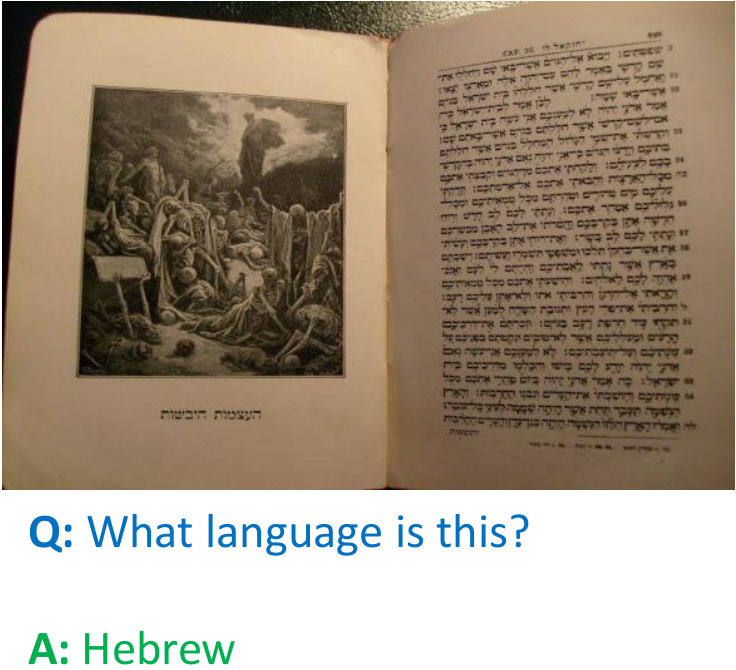}
    \label{fig:exist_c}}\hspace{0.05cm}
    \subfigure[]{\includegraphics[width=3.5cm,height=3.2cm]{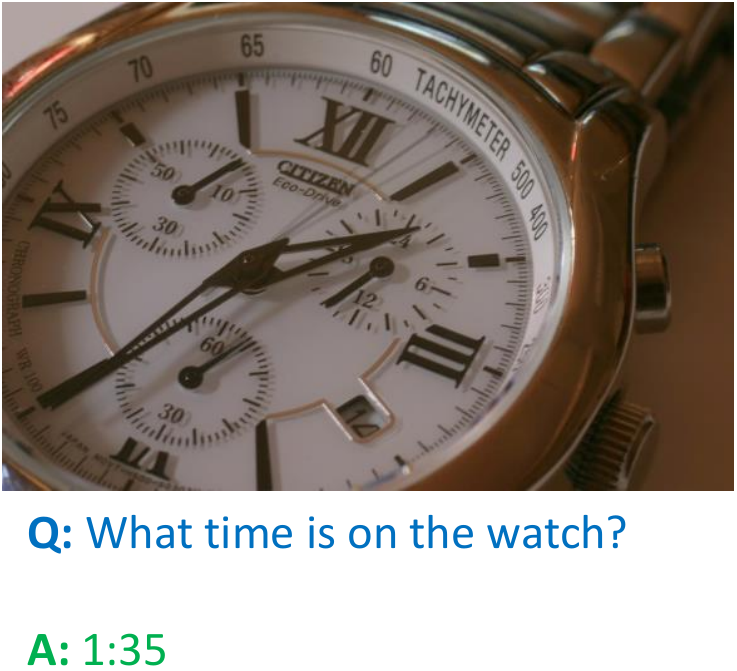}
    \label{fig:exist_d}}
    \caption{Some example images and QA pairs from the Text-VQA proposed in~\cite{singh2019towards}. Four different types of issue are shown. (a) questions that can be answered without reading image text; (b) questions that have more than one correct answer; (c) questions that require a large amount of external knowledge to answer; (d) questions that require skills that cannot be learned from the training data alone.}
    \label{fig:existing_datasets}
\end{figure*}

\begin{figure*}[t!]
    \centering
    \includegraphics[width=0.3\textwidth]{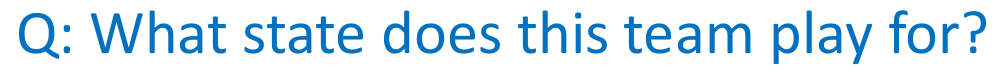}

    \subfigure[Original Image]{\includegraphics[width=3.5cm,height=2.3cm]{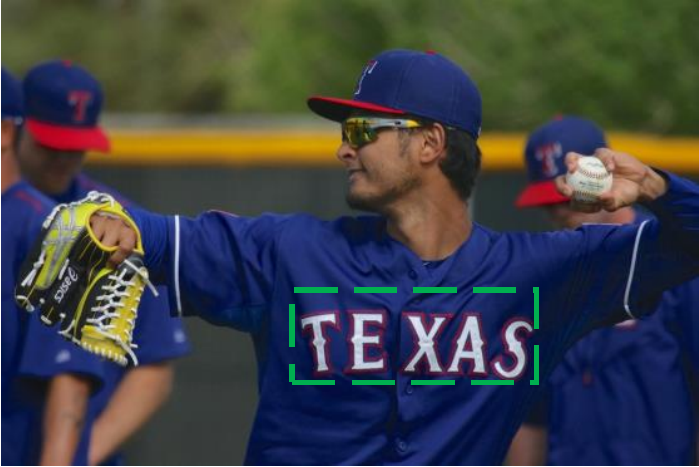}\label{fig:unreasonable_a}}
    \hspace{0.05cm}
    \subfigure[Cropped Image w/ text]{\includegraphics[width=3.5cm,height=2.3cm]{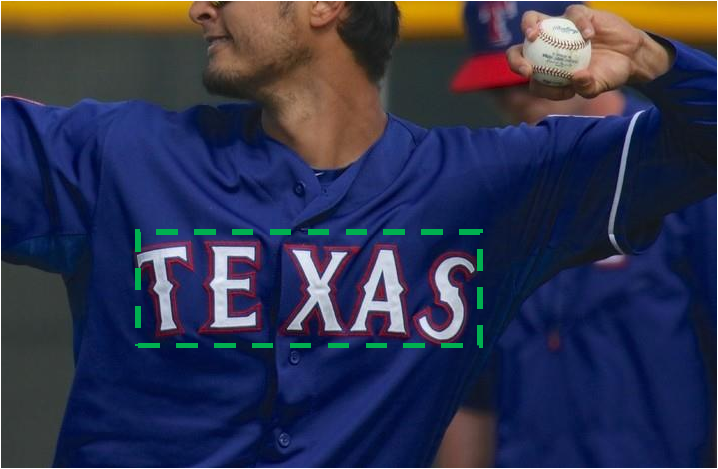}\label{fig:unreasonable_b}}
    \hspace{0.05cm}
    \subfigure[Cropped Image w/o text]{\includegraphics[width=3.5cm,height=2.3cm]{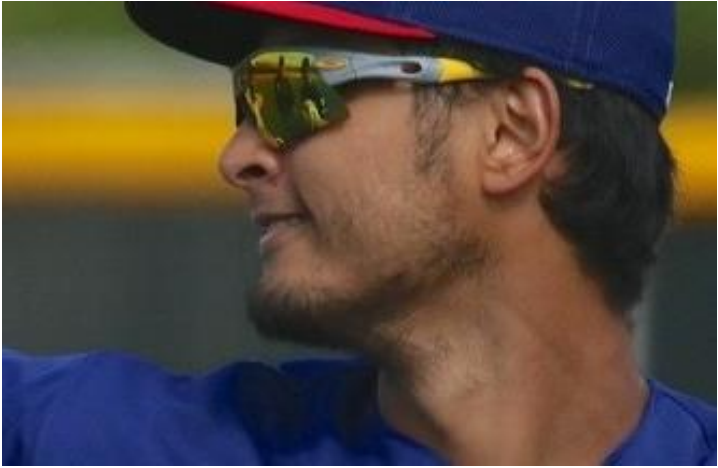}\label{fig:unreasonable_c}}
    \hspace{0.05cm}
    \subfigure[Occluded Image]{\includegraphics[width=3.5cm,height=2.3cm]{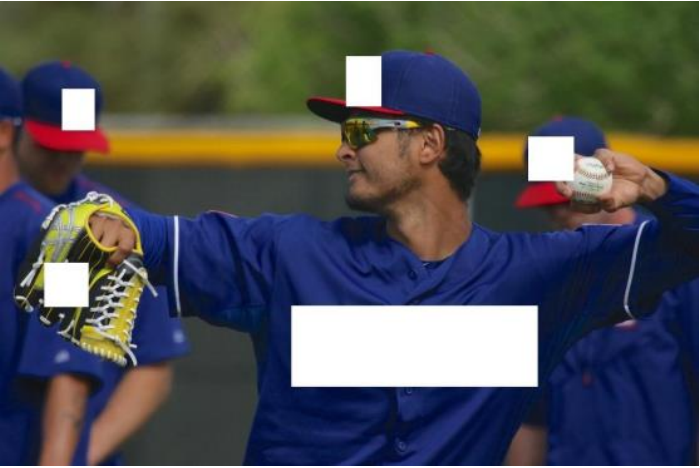}\label{fig:unreasonable_d}}

    \includegraphics[width=0.82\textwidth]{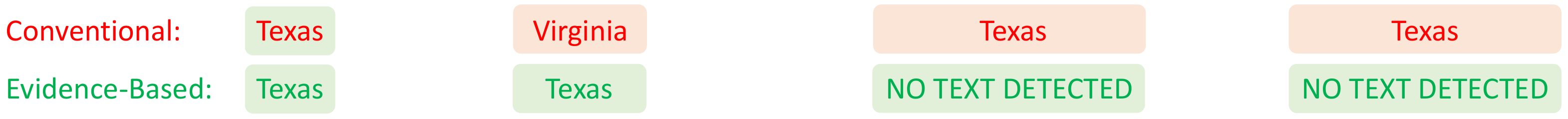}
    \caption{A comparison of conventional (LoRRA \cite{singh2019towards}), and evidence-based VQA methods.}
    \label{fig:unreasonable}
\end{figure*}

The version of the VQA problem that we apply this approach to is Scene Text VQA.
Several recent works \cite{biten2019scene, singh2019towards} have revealed that current VQA models perform badly on text VQA datasets, so it represents a compelling challenge falling within the existing framework. The various forms of text VQA problem are also of great practical import, because text represents a critical cue to understanding the content of an image. More than this, text VQA problems are typically less susceptible to solving through exploiting coincidental correlations in the data.

A variety of text-based VQA datasets~\cite{biten2019scene, kembhavi2017you, mishraocr, singh2019towards} have been proposed.
However, there is still a significant gap between current algorithm performance and that required to support practical applications~\cite{biten2019scene, mishraocr, singh2019towards}.
Another motivating factor in selecting text-based VQA rather than the generic version of the problem is that the text-based version of the problem is less susceptible to n-way classification over a fixed vocabulary. This is due to the fact that the range of text appearing in images is quite broad. The classification-based approach has repeatedly been shown to be susceptible to overfitting~\cite{agrawal2018don,goyal2017making}. Text-based VQA requires the development of alternative approaches, some of which will hopefully generalize.

Fig.~\ref{fig:existing_datasets} depicts some of the challenges with existing scene-text based VQA system. For example, Fig.~\ref{fig:exist_a} is a sample question that can be answered without reference to any textual content; while the question in Fig.~\ref{fig:exist_b} could have more than one correct answer; the question in Fig.~\ref{fig:exist_c} requires prior knowledge to answer; and finally in Fig.~\ref{fig:exist_d}, the answer can not be obtained directly from the text in the image, but require other skills. Such questions are contrary to the aims of text-based VQA and introduce bias into the dataset.

Empirical results presented below demonstrate that current VQA approaches rely heavily on a pre-defined answer space constructed by analysis of the answers in the training set, and limiting generalization. As shown in Fig.~\ref{fig:unreasonable_b}, their dependence on superficial image features can render conventional VQA methods sensitive to image modifications that do not change the semantics. Fig.~\ref{fig:unreasonable_c} and \ref{fig:unreasonable_d} demonstrate their propensity to generate an answer even when the required information is not present.
Text-VQA \cite{singh2019towards} employed the generic VQA accuracy as the performance metric, while ST-VQA \cite{biten2019scene} used a soft score metric inspired by the optical character recognition community. Both of these metrics are results-oriented, which means that a prediction is correct if it is identical to the ground-truth. They do not assess the reasoning process.
This leads to classification-based VQA models that are prone to overfit a fixed answer space. This enables impressive performance, but poor generalized to other datasets.
To address these issues, we propose a new scene-text based VQA dataset called `Scene Text+Evidence Visual Question Answering' (STE-VQA). Based on this, three tasks namely \textit{cross language challenge}, \textit{localization challenge} and \textit{traditional challenge} are introduced to motivate the creation of solutions with practical value from various aspects. Also, a series of baseline experiments were conducted to establish a lower bound for the three challenges. The main contributions of this paper are outlined as follows:
\begin{itemize}
\setlength{\itemsep}{-0.12cm}
    \item \textbf{Dataset:} The STE-VQA dataset provides questions, images and answers, but also a bounding box for each question that indicates the area of the image that informs the answer. We refer to such bounding boxes as \emph{evidence}.
    The dataset is intended to enable the development of text VQA methods that are closer to the levels of performance required by practical applications, but also to encourage the development of general VQA Methods that generalize.
    \item \textbf{Evaluation Metric:} We introduce an Evidence-based Evaluation (EvE) metric, which will require a VQA model to provide evidence to support the predicted answer. For this purpose, a new VQA model is also proposed. Under this new metric, it is anticipated that it will be much more difficult for naive classification models to achieve inflated performance.
    \item \textbf{Bilingual:} To the best of our knowledge, the proposed STE-VQA is the first bilingual scene text VQA dataset that includes both English and Chinese question and answer pairs. The fact that the proposed dataset embodies questions in two languages further rewards methods that generalize well. It is more difficult for a method to exploit superficial correlations in questions expressed in multiple languages. The languages chosen are also particularly grammatically distinct, and reflect culturally distinct populations, which leads to different question statistics, and further encourages generalization.
\end{itemize}

\subsection{Related Work}

Visaul Question Answering has gained significant attention recently, partly because it seems so unlikely that a method might be capable of answering all possible questions about all possible images\cite{antol2015vqa, malinowski2014multi}. Readers are encouraged to refer to \cite{kafle2017visual, wu2017visual} for a complete overview. Due to space constraint, this section only reviews the most relevant works to this paper, \emph{i.e.}, text-based VQA.

\begin{figure}[t!]
    \centering
    \includegraphics[width=0.47\textwidth]{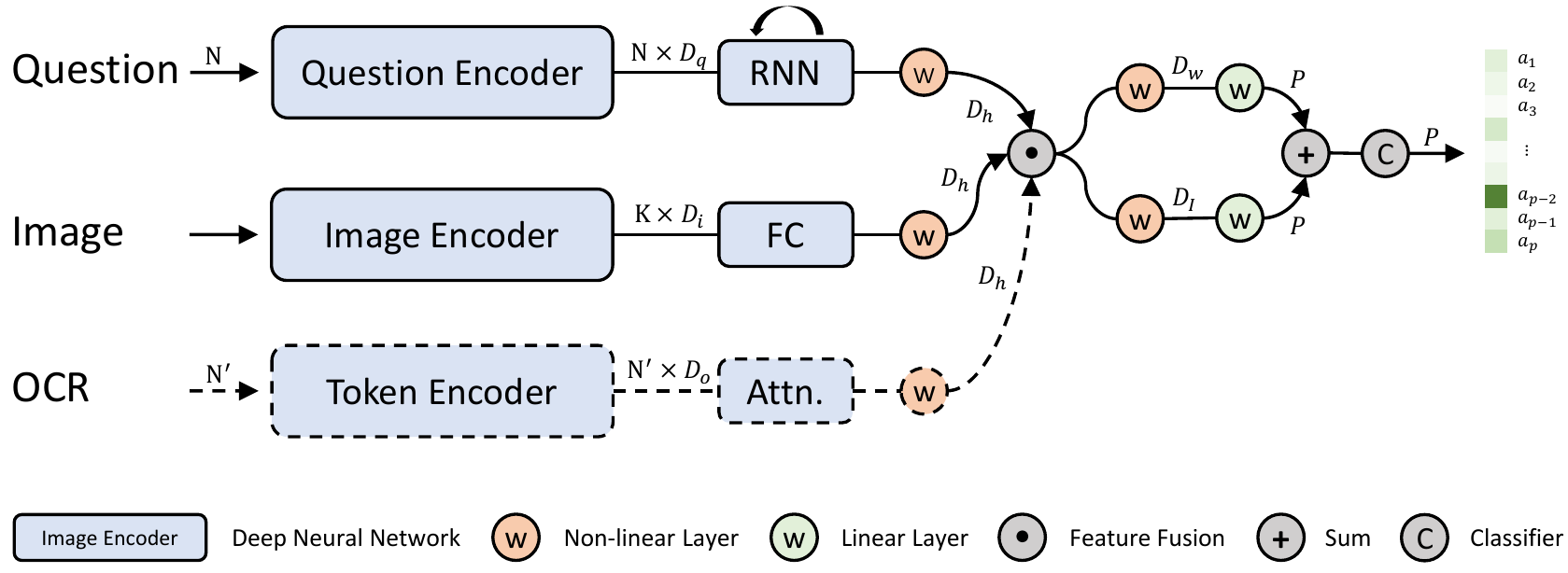}
    \caption{Illustration of the mainstream VQA models. $D_{q}$, $D_{i}$, $D_{o}$ and $D_{h}$ are the dimensions of the word embedding, image feature, OCR token embedding and hidden vector representations respectively. $N$, $N^{'}$ and $P$ indicate question length, number of OCR tokens and answer space. Blocks with dashed lines are optional modules used for text-based VQA.}
    \label{fig:mainstream_framework}
    \vspace{-0.5cm}
\end{figure}

\subsection{Text-based VQA}

In contrast to generic VQA datasets \cite{kafle2017visual, wu2017visual}, text-based VQA datasets pay more attention to text related questions where a VQA model is required to read and understand textual content in an image. In \cite{singh2019towards}, the authors proposed a dataset and baseline model, called Text-VQA and LoRRA respectively. LoRRA follows the structure of mainstream VQA models (see Fig.~\ref{fig:mainstream_framework}) where image features and word embedding are fused to train a classifier. Later, two other similar datasets were introduced, \emph{i.e.}, ST-VQA \cite{biten2019scene} and OCR-VQA \cite{mishraocr}. All these three datasets provide images with text related question and answer pairs. However, there are several important differences between them, as well as to our proposed dataset:

\begin{figure*}[t!]
    \centering
    \subfigure[English Question]{
        \includegraphics[width=0.5\textwidth]{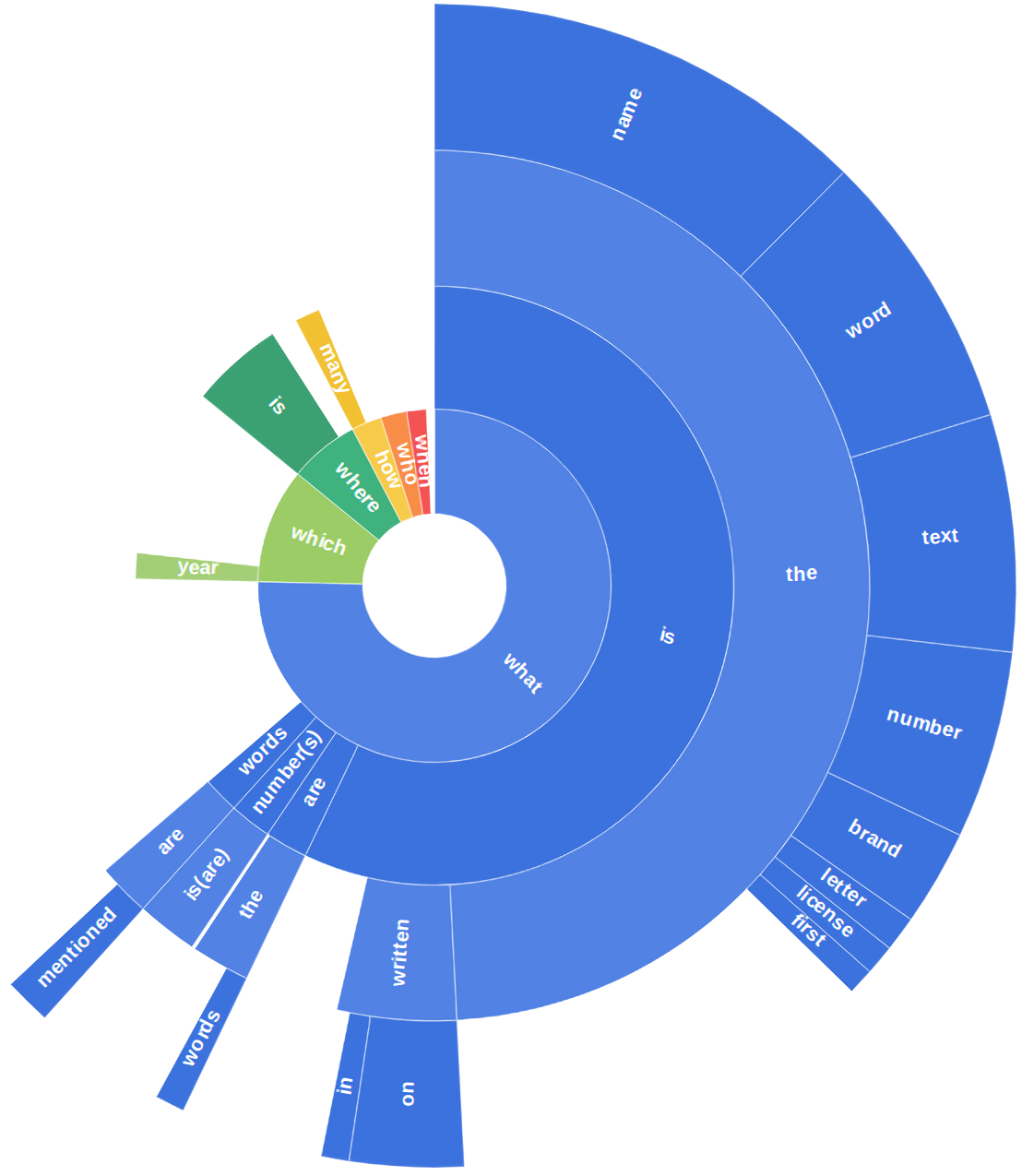}
    }
    \hspace{2.5cm}
    \subfigure[Chinese Question]{
        \includegraphics[width=0.5\textwidth]{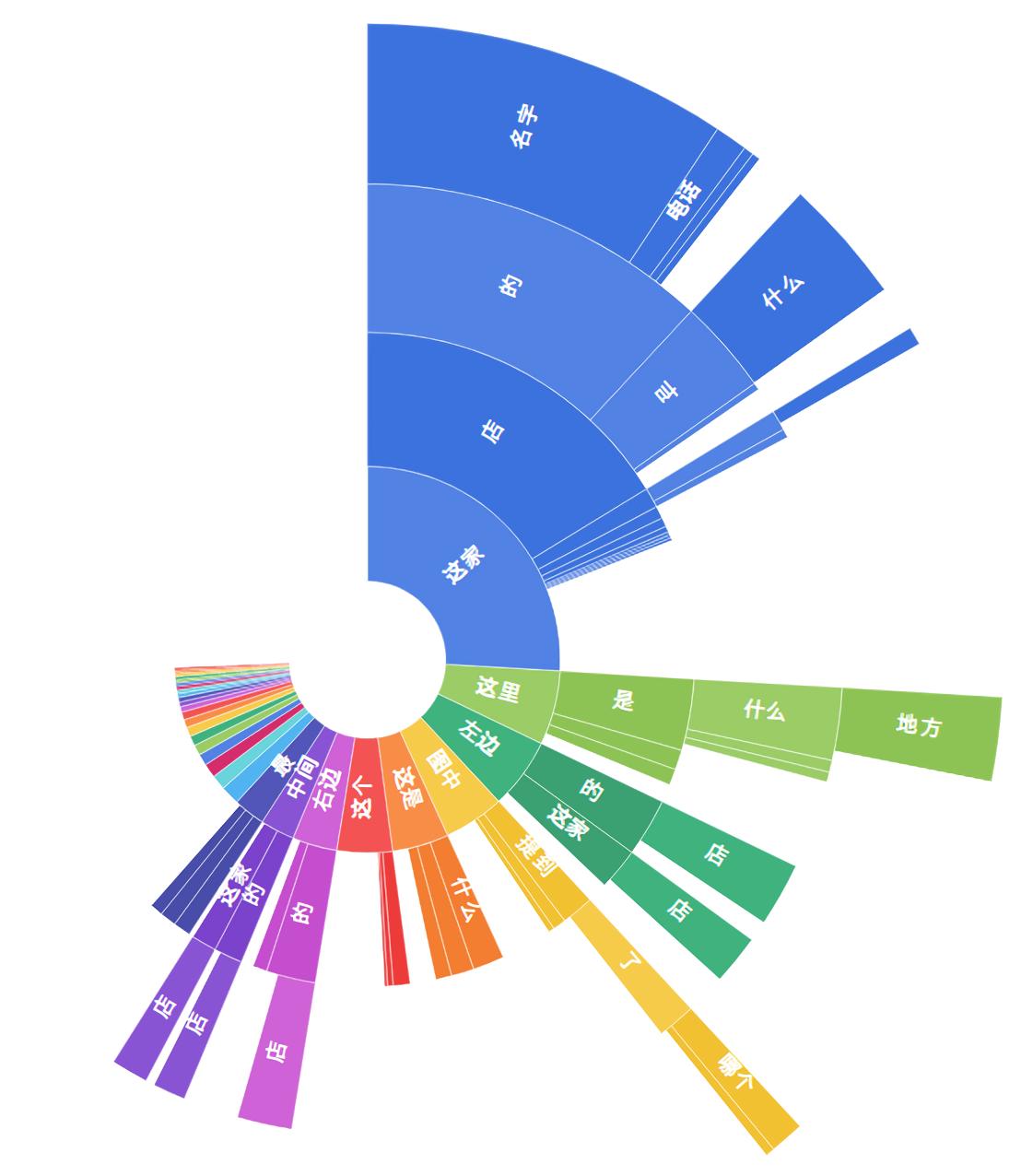}
    }
    \caption{Distribution of first four words in questions in STE-VQA.}
    \label{fig:que_distribution}
\end{figure*}

\noindent\textbf{Diversity:} Table~\ref{tab:diversity} shows the size and image sources of existing datasets and our dataset. Both of the Text-VQA \cite{singh2019towards} and OCR-VQA \cite{mishraocr} images came from a single image database which is Open Images v3 dataset \cite{openimages} and Book Cover Dataset \cite{iwana2016judging} respectively. While ST-VQA \cite{biten2019scene} was built upon a combination of public image datasets that include multiple tasks, \emph{e.g.}, text detection \cite{karatzas2015icdar, veit2016coco}, image classification \cite{deng2009imagenet}, generic visual question answering \cite{gurari2018vizwiz}, etc. It is noteworthy that although \cite{mishraocr} has the highest amount of images and QA pairs, the images are all of book covers, thus the diversity of images and questions are very limited. STE-VQA dataset stands out among other text VQA datasets with the consideration that existing datasets pay more attention to the question answering part, and the OCR part is almost ignored in both training and evaluation of the model.

\begin{table}[t!]
\footnotesize
\begin{center}\begin{tabular}{c|l|l|l|l|l}
\cline{1-6}
\multirow{2}{*}{{\scriptsize Dataset}} & \multicolumn{2}{c|}{{\scriptsize Train + Val}} & \multicolumn{2}{c|}{{\scriptsize Test}} & \multicolumn{1}{c}{\multirow{2}{*}{Image Source}} \\ \cline{2-5}
 & \multicolumn{1}{c|}{\# I} & \multicolumn{1}{c|}{\# Q} & \multicolumn{1}{c|}{\# I} & \multicolumn{1}{c|}{\# Q} & \multicolumn{1}{c}{} \\ \hline%
\cite{biten2019scene}    & 19k & 26k & 3k & \multicolumn{1}{l|}{4k} & \cite{deng2009imagenet, gurari2018vizwiz, karatzas2015icdar, karatzas2013icdar, krishna2017visual, mishra2013image, veit2016coco} \\
\cite{mishraocr}         & 180k & 900k & 20k & \multicolumn{1}{l|}{100k} & \cite{iwana2016judging} \\
\cite{singh2019towards}  & 25k & 39k & 3k & \multicolumn{1}{l|}{5k} & \cite{openimages}\\ \hline%
ours & 21k & 23k & 4k & 5k & \cite{ch2019total, karatzas2015icdar, karatzas2013icdar, liu2019curved, nayef2019icdar2019, sun2019chinese, veit2016coco} \\ \cline{1-6}
\end{tabular}\end{center}
\caption{A comparison of the amount and source of images between different text-based VQA datasets. \#I and \#Q indicate the number of images and questions respectively.}
\label{tab:diversity}
\vspace{-0.6cm}
\end{table}

\noindent\textbf{Evaluation Metric:} \cite{singh2019towards} employs a widely used VQA accuracy which was first proposed in \cite{goyal2017making}. Under this metric, each question has 10 answers that are labeled by different human annotators. Supposed that the prediction of a VQA model is $ans$, then the score for a single sample is calculated as:

\begin{equation}
    s_{v}(ans) = \min\{\frac{{\rm\#humans\ that\ said\ }ans}{3},\ 1\}
    \label{eqa:textVQA_metric}
\end{equation}

\noindent where \# indicates the number of human annotated labels that are identical to the predicted answer. This metric is robust against the incorrect answers given by some annotators. However, it is clear that only 4 discrete scores would appear, \emph{i.e.}, $\{0, \frac{1}{3}, \frac{2}{3}, 1\}$. In \cite{biten2019scene}, Levenshtein distance \cite{levenshtein1966binary} was proposed to softly penalize a mistake. Given the predicted answer $ans$ and ground-truth label $gt$, then the normalized Levenshtein similarity score $s_{l}$ is given as:

\begin{equation}
s_{l}(ans, gt) =
\begin{cases}
    \emph{$1 - NL(ans, gt)$},& \text{$NL(ans, gt) < \tau$} \\
    \emph{$0$},& \text{$NL(ans, gt) \geq \tau$} \\
\end{cases}
\label{eqa:stVQA_metric}
\end{equation}

\noindent where $\tau$ is a penalty threshold, and NL is the normalized Levenshtein distance between ground-truth and prediction.

\section{Proposed Dataset: STE-VQA}

A fundamental hypothesis in STE-VQA dataset is that a VQA model should answer a question correctly based on the textual content in an image. Therefore, we separate our scene text VQA tasks into two parts, \emph{i.e.}, 1) text spotting and 2) question answering. In this section, we describe the process to build the STE-VQA dataset. Also, we will detail the evidence-based evaluation metric and the new tasks for STE-VQA dataset.

\subsection{Data Collection}

\noindent\textbf{Images:} As STE-VQA dataset is designed for scene text VQA tasks, we collected a total of 20,757 images from publicly available scene text detection and recognition datasets. Specifically, images annotated with English questions and answers are obtained from Total-Text \cite{ch2019total}, ICDAR 2013 \cite{karatzas2013icdar}, ICDAR 2015 \cite{karatzas2015icdar}, CTW1500 \cite{liu2019curved}, MLT \cite{nayef2019icdar2019}, and COCO Text \cite{veit2016coco}. Whereas, images with Chinese questions and answers are collected from LSVT \cite{sun2019chinese}. All the images originated from these scene text datasets are comprised of daily scenes that include both indoor and outdoor settings.

\begin{figure}[t!]
    \centering
    \includegraphics[width=0.4\textwidth]{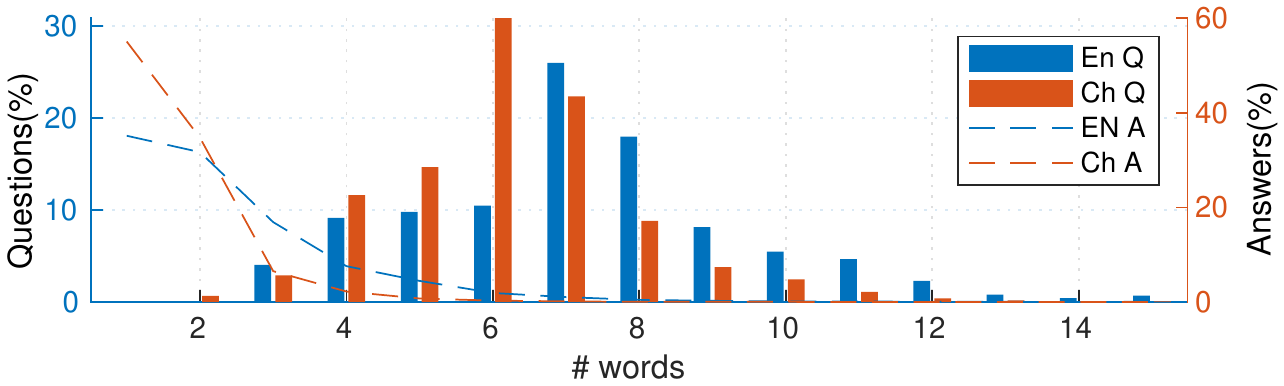}
    \caption{Percentage of question and answer length. Questions are tokenized by words. En and Ch stand for English and Chinese respectively.}
    \label{fig:qe_ans_length}
\end{figure}

\noindent\textbf{Questions and Answers:} The proposed STE-VQA dataset consists of 15,056 English questions and 13,006 Chinese questions. For the collection of question and answer pairs, annotators were requested to come up with questions that can be answered only by reading texts in the images. In order to avoid the question that does not require reading any text in the image, annotators are enforced to label a corresponding quadrilateral bounding box of the textual answer. The annotated bounding box will then serve as evidence to support the answer. Moreover, yes/no questions and ambiguous questions that could have multiple correct answers are prohibited. Fig.~\ref{fig:que_distribution} shows the common types of question, it is clear that most of the English questions start with ``what", %
and following with `is' and `the'.
However, the composition of Chinese questions is far more complex than the English questions due to different grammar, vocabulary and other characteristics of the Chinese language. Fig.~\ref{fig:qe_ans_length} shows the distribution of the length of questions and answers. Different from English words which can be segmented by space directly, Chinese words are composed of multiple Chinese characters in a continuous sentence. Therefore, we use \cite{sun2012jieba} to tokenize Chinese questions for counting the percentage of question length. From the Fig.~\ref{fig:qe_ans_length}, it is clear that most of the English and Chinese questions have between 6 to 8 words, and the majority of their answers are of a single word.

In summary, as shown in Table~\ref{tab:FT_VQA_Volume}, 25,239 images and 28,062 QA pairs are separated into 20,757 images 23,062 questions for training set and 4,482 images 5,000 questions for testing set.

\begin{table}[t!]
\footnotesize
\begin{center}
\begin{tabular}{l|cc|cc|cc}
\hline
\multicolumn{1}{c|}{\multirow{2}{*}{Set}} & \multicolumn{2}{c|}{English} & \multicolumn{2}{c}{Chinese} & \multicolumn{2}{|c}{All} \\ \cline{2-7}
\multicolumn{1}{c|}{} & \multicolumn{1}{c|}{\# I} & \multicolumn{1}{c|}{\# Q} & \multicolumn{1}{c|}{\# I} & \multicolumn{1}{c}{\# Q}  & \multicolumn{1}{|c|}{\# I} & \multicolumn{1}{c}{\# Q}\\ \hline%
Train & 11,383 & 12,556 & 9,374 & 10,506 & 20,757 & 23,062 \\ %
Test & 2,267 & 2,500 & 2,215 & 2,500 & 4,482 & 5,000 \\ %
Total & 13,650 & 15,056 & 11,589 & 13,006 & 25,239 & 28,062 \\ \hline
\end{tabular}
\end{center}
\caption{Volume of the STE-VQA dataset.}
\label{tab:FT_VQA_Volume}
\end{table}

\subsection{Evidence-based Evaluation (EvE) Metric}

We observed an intriguing trend among the classification based approaches for scene text VQA task. That is to say, if the ground-truth answer was included in the pre-generated answer dictionary, a generic VQA model may predict a correct answer without reading the textual content. However, such methods rely heavily on the pre-defined answer pool and so, they are unable to handle questions with out-of-vocabulary answers. Therefore, it is unclear whether such models truly have the capability to understand and reason about the questions or they are merely over-fitting to the fixed answer space. Inspired by this observation, we introduce a new evaluation protocol, named Evidence-based Evaluation (EvE) metric, which will require a VQA model to provide evidence to support the predicted answers. Under this metric, it will be much more difficult for naive classification models to achieve inflated performance.

\begin{figure}[t!]
    \centering
    \includegraphics[width=0.4\textwidth]{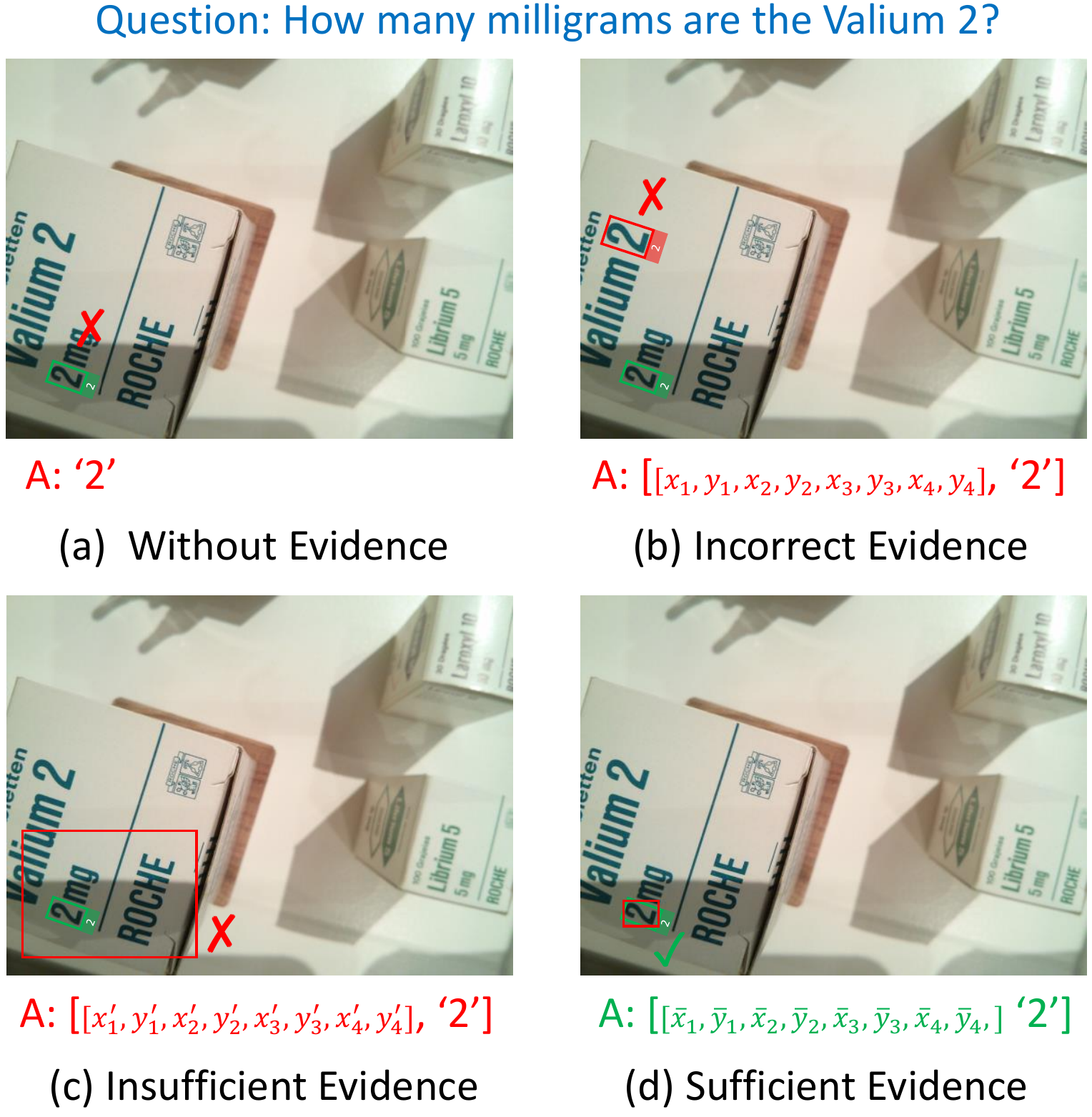}
    \caption{In EvE metric, evidence in the form of bounding box should be given as well as the predicted answer. Green and red bounding boxes are ground-truth and predicted evidence respectively. \textbf{Incorrect}: (a) answer without evidence; (b) answer with inappropriate evidence; (c) answer with insufficient evidence. \textbf{Correct}: (d) answer with appropriate evidence. It is worth mentioning that all of the above answers would be marked as correct in the conventional VQA evaluation metric because all of them give the right answer `2'.}
    \label{fig:evaluation_protocol}
    \vspace{-0.5cm}
\end{figure}

Generally, EvE metric consists of two steps: a) check the answer; b) check the evidence. In the former, we use the normalized Levenshtein similarity score (see Eq.~\eqref{eqa:stVQA_metric}). In the latter, we adopt the widely used IoU metric to determine whether the evidence is sufficient or insufficient. Suppose $B_{gt}$ and $B_{det}$ are the ground-truth and predicted bounding box respectively, then the evidence sufficiency score, $E$ is defined as:

\begin{equation}
E_{\tau}^{i} = f(\frac{B_{gt} \cap B_{det}}{B_{gt} \cup B_{det}}) =
    \begin{cases}
    \emph{Incorrect},& \text{$E = 0$}\\
    \emph{Insufficient},& \text{$0 < E < \theta$} \\
    \emph{Sufficient},& \text{$ E \geq \theta$}
    \end{cases}
    \label{eqa:evidence_sufficiency}
\end{equation}

\noindent where $\theta=0.5$ is an predefined threshold. Under the EvE metric, only \emph{correct} answers with \emph{sufficient} evidence contribute to the final performance $s_{e}$ (see Fig.~\ref{fig:evaluation_protocol}) where it is given by:

\begin{equation}
    s_{e}(ans, gt, E) =
    \begin{cases}
        \emph{$s_{l}$},& \text{if $E$ sufficient} \\
        \emph{$0$},& \text{else}
    \end{cases}
    \label{eqa:evidence_based_metric}
\end{equation}

\noindent where $s_{l}$ is the normalized Levenshtein similarity score as defined in Eq.~\eqref{eqa:stVQA_metric}.

\begin{figure*}[t!]
    \centering
    \includegraphics[width=0.8\textwidth]{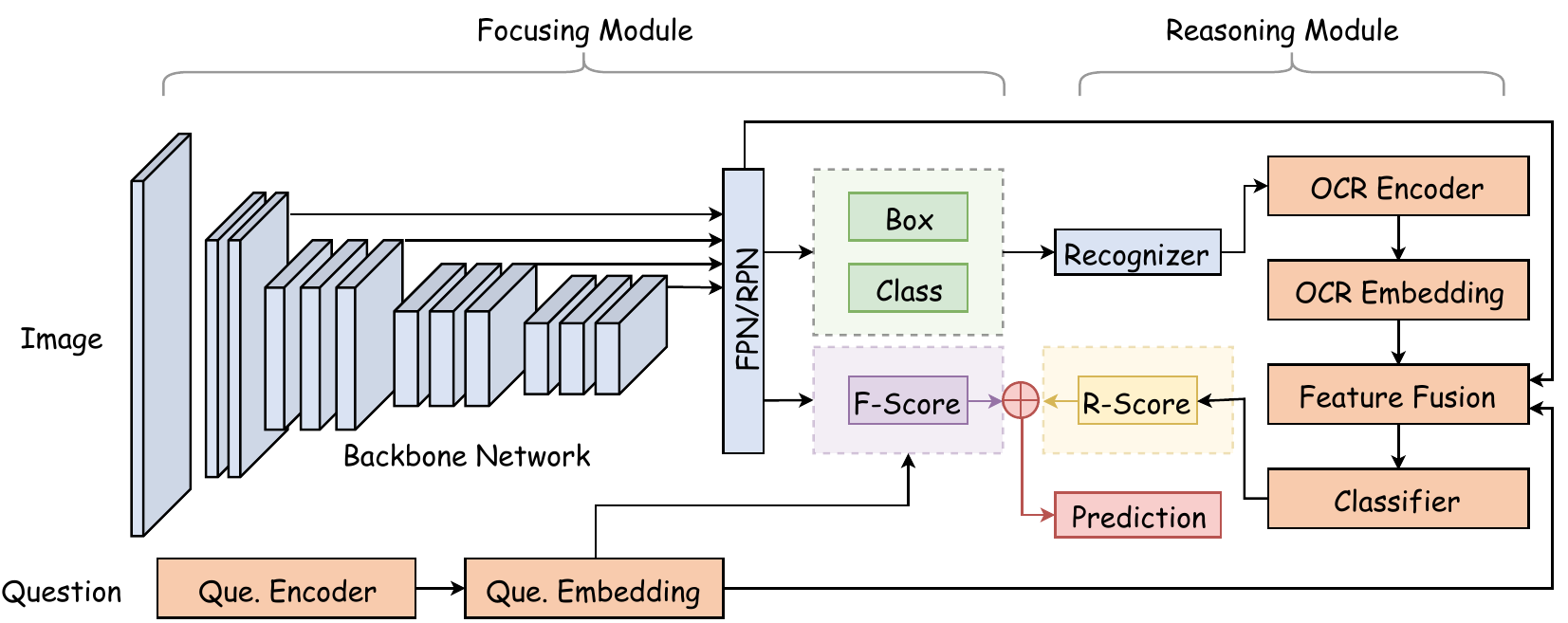}
    \caption{Overview of the QA R-CNN architecture}
    \label{fig:pipeline_qa_rcnn}
\end{figure*}

\subsection{Tasks}

Both Text-VQA \cite{singh2019towards} and OCR-VQA \cite{mishraocr} follow the same rules as presented in generic question answering task. Although ST-VQA \cite{biten2019scene} proposed three tasks, the only difference between each of the tasks is the size of external information (vocabulary), which is insignificant and unreasonable to properly evaluate the models' full capability. For instance, in the strongly contextualised task, all ground-truth answers are provided in a dictionary for every image with a set of distractors, which makes the VQA model prone to overfit the provided vocabulary. Besides, it becomes more difficult for these models that are trained on a fixed dictionary to generalize to other datasets.

As a result of this, we propose three related tasks namely as \textit{Cross Language Challenge}, \textit{Localization Challenge}, and \textit{Traditional Challenge} that will be detailed below to improve the task diversity. An online evaluation server will be set up for results submission.

\begin{itemize}
\setlength{\itemsep}{-0.12cm}
\item
\noindent\textbf{Cross Language Challenge (CLC):} As the proposed STE-VQA dataset is a bilingual VQA dataset that contains both English and Chinese QA pairs, this challenge aims to explore a model's ability in extracting the common knowledge between different languages. Under this challenge, the candidates are requested to submit results predicted by both the monolingual (\emph{English-only}, \emph{Chinese-only}) and bilingual models with an identical framework (\emph{e.g.} network structure) for evaluations. The proposed EvE metric is used to evaluate the model's performance in this challenge.

\item
\noindent\textbf{Localization Challenge (LC):} To gain insights of a VQA model, we encourage candidates to train an evidence based VQA model to simultaneously predict the answer and its corresponding bounding box as evidence, instead of simply employing an off-the-shelf OCR system to obtain the OCR tokens. Hence, the main objective of this challenge is to explore the VQA model's ability in understanding the question and locating the correct image space that contains the answers. That is to say, this challenge requires the VQA model to provide the spatial location where an answer will be most likely to appear in an image based on a question. Compared to the full challenge, LC ignores the text recognition error and the difficulties of combining multiple OCR tokens for long answers. IoU between the predicted and ground-truth bounding box is employed as the performance metric for this challenge.

\item
\noindent\textbf{Traditional Challenge (TC):} We maintain the traditional VQA challenge that is consistent with the existing VQA datasets in which this challenge does not consider the evidence for the predicted answers. The normalized Levenshtein similarity score between the prediction and ground-truth is employed as the metric for this challenge.
\end{itemize}

\begin{table*}[t!]
\footnotesize
\begin{center}
\begin{tabular}{l|cc|ccccc|ccc|cc|ccc|c}
\hline
\multicolumn{1}{c|}{\multirow{3}{*}{Model}} & \multicolumn{7}{c|}{CLC (\%)} & \multicolumn{3}{c|}{LC (\%)}  & \multicolumn{5}{c|}{TC (\%)} & $\Delta_{r}$ \\ \cline{2-17}
\multicolumn{1}{c|}{} & \multicolumn{2}{c|}{Mono.} & \multicolumn{5}{c|}{Bi.} & \multicolumn{3}{c|}{Bi.} & \multicolumn{2}{c|}{Mono.} & \multicolumn{3}{c|}{Bi.} \\
\multicolumn{1}{c|}{} & En & Ch & En & Ch & \multicolumn{1}{c}{S} & \multicolumn{1}{c}{L} & Acc & En & Ch & Acc & En & Ch & En & Ch & Acc \\ \cline{1-17}
SV UB & - & - & - & - & - & - & - & - & - & - & 31.1 & 7.8 & 31.3 & 8.9 & 20.1 & - \\
LV UB & - & - & - & - & - & - & - & - & - & - & 48.0 & 16.1 & 48.3 & 17.0 & 32.7 & - \\
OCR UB & 33.9 & 24.5 & 33.9 & 24.5 & 44.1 & 14.3 & 29.2 & 50.0 & 37.8 & 43.9 & 38.5 & 28.2 & 38.5 & 28.2 & 33.3 & - \\ \hline
Random & 4.4 & 1.1 & 4.7 & 1.2 & 5.1 & 0.8 & 3.0 & 15.1 & 5.1 & 10.1 & 5.8 & 1.5 & 5.9 & 1.5 & 3.7 & 0.81 \\
P\cite{singh2018pythia}+SV & 4.3 & 0.1 & 4.5 & 0.1 & 4.3 & 0.2 & 2.3 & 17.2 & 1.8 & 9.5 & 8.0 & 0.7 & 7.7 & 0.7 & 4.2 & 0.54 \\
P\cite{singh2018pythia}+LV & 4.7 & 0.2 & 4.4 & 0.2 & 4.2 & 0.3 & 2.3 & 17.4 & 2.4 & 9.9 & 9.2 & 0.8 & 8.2 & 0.6 & 4.4 & 0.52\\
L\cite{singh2019towards}+SV & \color{red}{8.2} & 1.2 & 8.4 & 2.0 & 9.6 & 0.8 & 5.2 & 18.0 & 5.4 & 11.7 & \color{red}{12.0} & 2.6 & \color{red}{13.2} & 3.3 & 8.2 & 0.63\\
L\cite{singh2019towards}+LV & 7.7 & 0.5 & 6.8 & 0.7 & 6.8 & 0.7 & 3.8 & \color{red}{18.5} & 3.9 & 11.2 & \color{red}{12.0} & 1.6 & 11.2 & 1.7 & 6.5 & 0.58\\ \hline
{\scriptsize QA R-CNN} & \cellcolor[HTML]{EEEEEE}7.7 & \cellcolor[HTML]{EEEEEE}1.4 & \cellcolor[HTML]{EEEEEE}\color{red}{8.8} & \cellcolor[HTML]{EEEEEE}\color{red}{3.2} & \cellcolor[HTML]{EEEEEE}\color{red}{10.8} & \cellcolor[HTML]{EEEEEE}\color{red}{1.1} & \cellcolor[HTML]{EEEEEE}\color{red}{6.0} & \cellcolor[HTML]{EEEEEE}18.3 & \cellcolor[HTML]{EEEEEE}\color{red}{7.3} &  \cellcolor[HTML]{EEEEEE}\color{red}{12.8} & \cellcolor[HTML]{EEEEEE}9.6 & \cellcolor[HTML]{EEEEEE}2.2 & \cellcolor[HTML]{EEEEEE}10.6 & \cellcolor[HTML]{EEEEEE}4.0 & \cellcolor[HTML]{EEEEEE}7.3 & \cellcolor[HTML]{EEEEEE}\color{red}{0.82} \\
{\scriptsize QA R-CNN w/ tricks}& \cellcolor[HTML]{EEEEEE}7.4 & \cellcolor[HTML]{EEEEEE}\color{red}{1.5} & \cellcolor[HTML]{EEEEEE}8.4 & \cellcolor[HTML]{EEEEEE}2.9 & \cellcolor[HTML]{EEEEEE}10.3 & \cellcolor[HTML]{EEEEEE}1.0 & \cellcolor[HTML]{EEEEEE}5.7 & \cellcolor[HTML]{EEEEEE}18.3 & \cellcolor[HTML]{EEEEEE}7.2 &  \cellcolor[HTML]{EEEEEE}\color{red}{12.8} & \cellcolor[HTML]{EEEEEE}11.8 & \cellcolor[HTML]{EEEEEE}\color{red}{7.9} & \cellcolor[HTML]{EEEEEE}12.7 & \cellcolor[HTML]{EEEEEE}\color{red}{9.4} & \cellcolor[HTML]{EEEEEE}\color{red}{11.0} & \cellcolor[HTML]{EEEEEE}0.52\\ \hline
\end{tabular}
\caption{Quantitative results of the three tasks in STE-VQA dataset. Mono. and Bi. represent monolingual and bilingual model respectively while S and L are short (one word) and long (more than one word) answers.}
\label{tab:three_tasks_performance}
\end{center}
\vspace{-0.45cm}
\end{table*}

\section{Baselines and Results}

\subsection{Baseline Methods}

This section presents the naive baseline models and two state-of-the-art VQA methods \cite{singh2018pythia, singh2019towards} that were employed in the experiments. This helps to show the difficulty of the proposed STE-VQA dataset and the new tasks. The entire STE-VQA dataset is separated into \emph{training} and \emph{testing} sets (see Table~\ref{tab:FT_VQA_Volume}), and 10\% data from the \emph{training} set is used for validation.

\noindent\textbf{Vocabulary Upper Bound:} As both \cite{singh2018pythia} and \cite{singh2019towards} are classification based method, two dictionaries are built under the widely used rules. Specifically, a small vocabulary (SV) is built with 927 English and 365 Chinese answers that appeared more than once in the training set and a Larger Vocabulary (LV) is built with 8,102 English and 8,212 Chinese unique answers. %
We explore the upper bound accuracy of the pre-generated SV and LV. We assume that answers included in the dictionaries can always be predicted correctly with perfect evidence to calculate the upper bound accuracy.

\noindent\textbf{OCR Upper Bound:} Since the traditional VQA models cannot obtain OCR tokens and info directly, we employ the state-of-the-art pre-trained text detection and recognition models \cite{liu2019omnidirectional,shi2016end} to extract OCR bounding boxes and characters. To evaluate the effectiveness of the OCR system, we calculate the OCR upper bound accuracy on the test set. All of the answers and evidence are directly obtained from the OCR results (and suppose the correct one can always be selected), it also considers combinations of up to 4 OCR tokens for multi-word answers.

\noindent\textbf{Random OCR Tokens:} To assess the arbitrary chance, this baseline returns a random OCR token and its bounding box from the OCR results for each question to obtain the random accuracy.

\noindent\textbf{State-of-the-art Approaches:} Both state-of-the-art generic \cite{singh2018pythia} and scene text \cite{singh2019towards} VQA models are employed as baselines to verify the difficulties of the STE-VQA dataset. It is important to note that these methods cannot provide evidence to support their predicted answers. Therefore, we queried the predicted answers from OCR results, \emph{i.e.}, if there are any identical OCR tokens to the predicted answer, then one of the predicted bounding boxes would be randomly selected as evidence, otherwise bounding box of the token which has the smallest normalized Levenshtein distance is selected.

\begin{figure}[t!]
    \centering
    \includegraphics[width=0.45\textwidth]{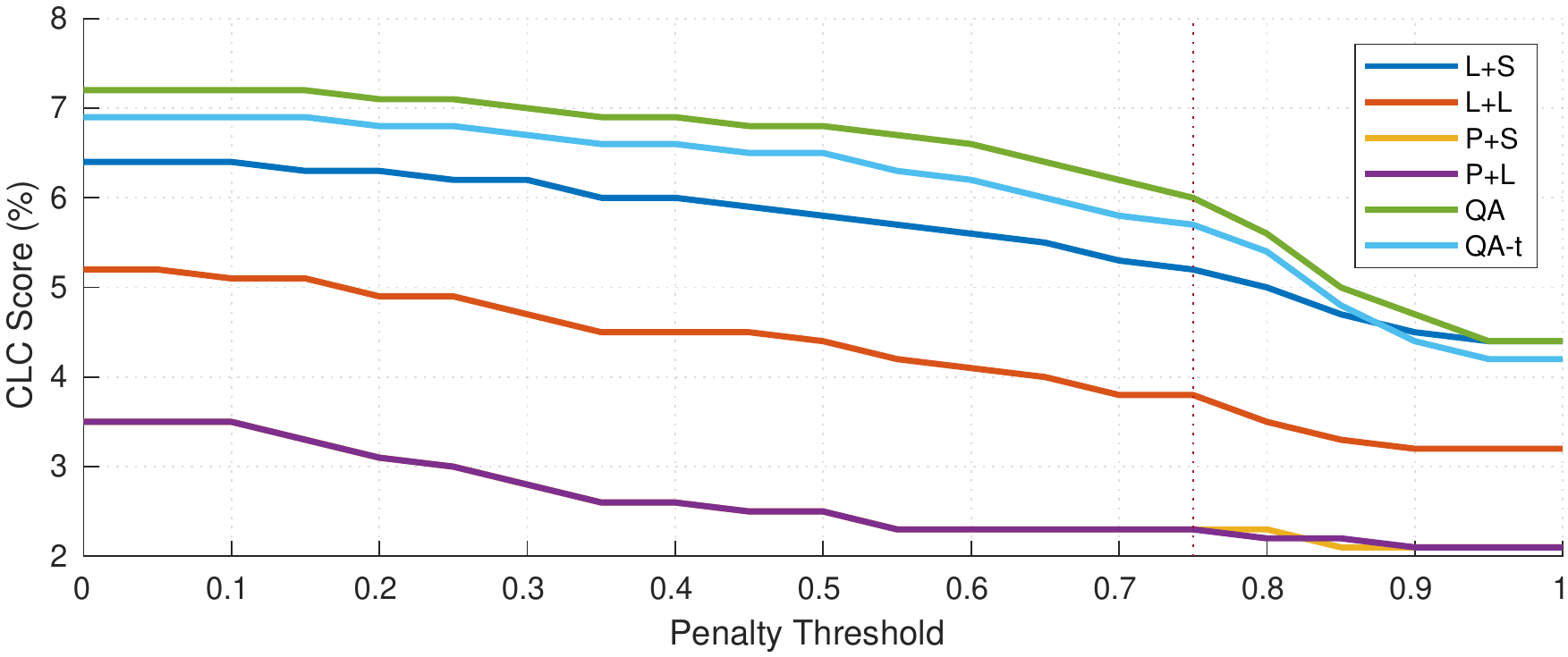}
    \caption{CLC score under different $\tau$}
    \label{fig:tau_effect}
    \vspace{-0.6cm}
\end{figure}

\noindent\textbf{QA R-CNN:} It is noteworthy that all of the aforementioned baseline methods cannot simultaneously output the answer and its corresponding bounding box as evidence. Therefore, we propose QA R-CNN. Generally, QA R-CNN consists of two parts: Focusing Module (FM) and Reasoning Module (RM) (see Fig.~\ref{fig:pipeline_qa_rcnn}). The core component in FM is a customized Faster R-CNN network trained for text detection task. Compared to the regular Faster R-CNN which only predicts bounding box and object category, QA R-CNN also outputs a focusing score for each bounding boxes. Technically, word embedding of question is first extracted by GloVe \cite{pennington2014glove} for English questions and Word2Vec \cite{li2018analogical} for Chinese questions. Then, the embedding is fed into LSTM layers to obtain question features. Following this, both question and image features are concatenated to classify the bounding box into answer area and non-answer area. This enables the QA R-CNN to gain the ability to draw its attention to the area that the answer may appear in the image. As such, a straightforward idea is that the model can directly use the underlying text of the bounding box with the highest focusing score as the question's answer. However, this choice will not consider the rich semantics of the textual content. Therefore, RM is introduced to further improve the pipeline. In RM, we follow the similar architecture in LoRRA where the semantics of detected text are further explored. Specifically, word embedding of the OCR tokens is extracted by FastText \cite{joulin2016bag} models that are pre-trained on English/Chinese Wikipedia, and then the OCR embedding is fused with both image features and question embedding for further classification. Different from other classification-based approaches, we do not use a pre-defined fixed dictionaries as the answer space, but only use the detected OCR tokens, \emph{i.e.}, only the detected text can be used as the answer. In the end, the weighted score of FM and RM are summed up for the final prediction.

\subsection{Results}

\begin{figure*}[t!]
    \centering
    \includegraphics[width=0.99\textwidth]{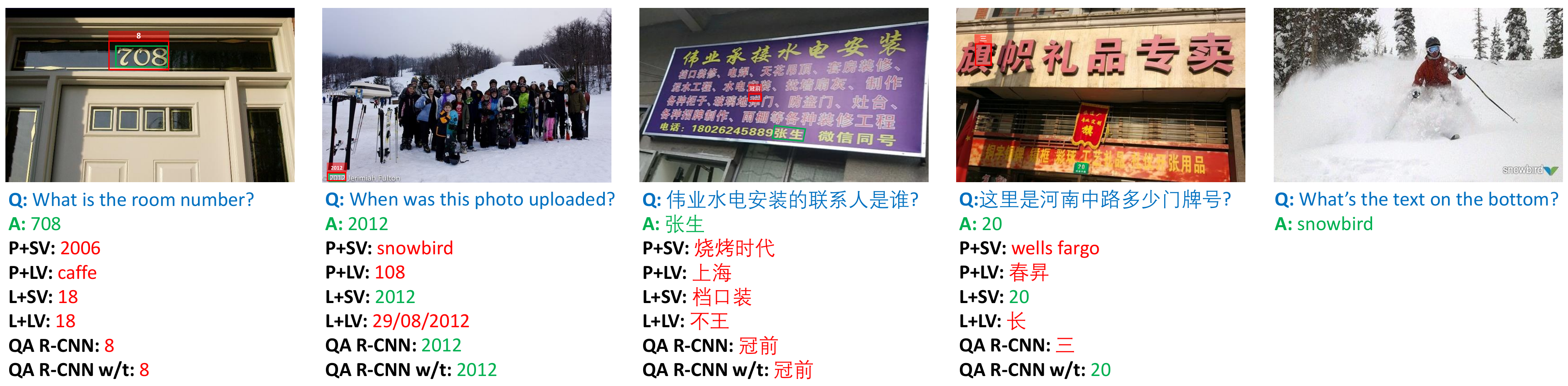}
    \caption{Visualization of the output answers on the STE-VQA dataset from different models (first four images). Green and Red bounding box are ground-truth and predicted evidence by QA R-CNN respectively. (More examples can be found in supplementary)}
    \label{fig:qualitative_res}
    \vspace{-0.35cm}
\end{figure*}

\noindent\textbf{Quantitative Results:} Table~\ref{tab:three_tasks_performance} summarizes the results of the baselines and our method on the STE-VQA dataset. The penalty threshold $\tau$ is practically set to 0.75 during the evaluation to ensure the answer quality. Fig.~\ref{fig:tau_effect} shows the CLC score under different $\tau$ for bilingual models.

We first measure the upper bound performance of the two pre-defined dictionaries SV and LV. Similar to other scene text VQA datasets, SV and LV can achieve high accuracy on English questions, \emph{i.e.}, 31.1 and 48.0 respectively. However, they failed catastrophically on the Chinese questions due to the language features and lower overlapping of answers between the training and testing splits. Hence, it is more difficult for the classification based method to obtain a promising performance on the Chinese split in the STE-VQA dataset. We also provide the upper bound accuracy of the OCR results that are generated by \cite{liu2019omnidirectional, shi2016end}, and it achieves better accuracy on Chinese questions compared to the fixed vocabularies. Then a baseline using random OCR token is set as a comparison with other approaches, and this heuristic method only achieves 3.0 and 3.7 overall score for the CLC and TC tasks respectively.

To further justify the need of STE-VQA, we trained two state-of-the-art approaches, \emph{i.e.}, Pythia (P) \cite{singh2018pythia} and LoRRA (L) \cite{singh2019towards}. As shown in Table~\ref{tab:three_tasks_performance}, Large Vocabulary (LV) helps both P and L to achieve better accuracy on English questions. However, both methods perform badly on Chinese questions due to a large amount of out-of-vocabulary answers in the test set. Also, as the CLC task requires a model to provide evidence as well as the answer, the accuracy of all of the studied methods dropped significantly as compared to the TC score. This is because the models infer the answers without actually reading the textual content in the images (see Fig.~\ref{fig:unreasonable_c} and \ref{fig:unreasonable_d}), thus they can not provide reasonable evidence to support the answer. In contrast, the proposed QA R-CNN shows more robust results on the three tasks (see Table~\ref{tab:three_tasks_performance}).

To further explore the proposed CLC task, we also trained a QA R-CNN with bells and whistles, many heuristic manual rules are adopted to lift the performance. Although this heavy model achieves top performance on the TC task, its CLC score is even lower than the baseline QA R-CNN. Such a scenario suggests that the evaluation protocol used in the current conventional VQA task is not reasonable to some extents, because the VQA models can easily overfit to the answer space by using tricks. Therefore, we introduce a {\it reasonable score} $\Delta_{r}$ to measure the percentage of answers with sufficient evidence, and it is denoted as $\Delta_{r} = \frac{{\rm CLC}_{all}}{{\rm TC}_{all}}$. Lower $\Delta_{r}$ means that the model has outputted many unreasonable but correct answers, which suggests that it might either overfit to the answer pool or use too many manual rules to achieve a higher score under conventional evaluation protocol. As shown in Table~\ref{tab:three_tasks_performance}, the QA R-CNN w/ tricks obtained the lowest reasonable score although it outperforms all other models under the traditional evaluation protocol. Another interesting observation is we found that all methods achieve extremely low accuracy on the questions that have a longer answer. We believe this is because current models cannot combine multiple texts together to generate a long answer. However, how to solve this issue is out of the scope of this paper, and thus we leave it for future work.

\noindent\textbf{Qualitative Results:} Fig.~\ref{fig:qualitative_res} illustrates some selected visualization results of the baseline methods. Surprisingly, we found that some models do not learn the concept of question type at all. For example, the `P+LV' model outputs a word `caffe' for the question `What is the room number?' that asks for a number, and `L+LV' predicts a character \begin{CJK*}{UTF8}{gbsn}`长' (long)\end{CJK*} for the question \begin{CJK*}{UTF8}{gbsn}`这里是河南中路多少门牌号' (What is the house number of this shop here in Henan Middle Road?)\end{CJK*} that asks for a number. Also, the incorrect recognition result is another reason that causes the incorrect answer, in the first sample shown in Fig.~\ref{fig:qualitative_res}, although correct bounding box of the answer `708' was predicted, it was however recognized as `8' and was further outputted as the answer. An interesting case is the `L+LV' model answers the question `When was this photo uploaded?' with `29/08/2012' when only `2012' appeared in the original image. Such a phenomenon tells us that similar answers in the vocabulary could interfere with the decision of classifier. Another noteworthy example is that `P+SV' model predicts `snowbird' for the question `When was this photo uploaded'. We queried another image with the answer `snowbird' in the training set (see last image in Fig.~\ref{fig:qualitative_res}) and it shows that the `P+SV' model outputs the same answer when the image contains similar visual features. Therefore, we believe that this VQA model might rely too heavily on the image feature and learned to map the image feature with the answer space but it does not truly understand the question. Additionally, for the question that requires stronger reasoning ability and image with many texts, such as the third sample in Fig.~\ref{fig:qualitative_res}, \begin{CJK*}{UTF8}{gbsn}`伟业水电安装的联系人是谁? (Who is the contact person for Weiye Hydropower Installation?)'\end{CJK*}, none of the models predict the answer correctly.

\section{Conclusion}

We have described a new bilingual scene text+evidence VQA dataset named STE-VQA that is annotated with both English and Chinese QA pairs.  Three related challenges are proposed, namely \textit{Cross Language}, \textit{Localization} and \textit{Traditional} that are designed to evaluate the generalization of VQA models. An evidence-based measure of an algorithm's capacity to reason is also proposed that requires the VQA model to provide a bounding box of the predicted answer. This metric aims to uncover whether the VQA model learns deeper relationships between text and image content, rather than overfitting to a pre-defined dictionary.
Future work includes extending the proposed EvE metric to existing VQA datasets in the hope that it might improve generalization and thus the practicality of VQA technologies.

{\small
\bibliographystyle{ieee_fullname}
\bibliography{ref}
}

\end{document}